\documentclass[10pt,twocolumn,letterpaper]{article}

\usepackage[pagenumbers]{cvpr}      
\usepackage{multirow}
\usepackage{multicol}

\usepackage{algorithm}
\usepackage{algpseudocode}

\usepackage{marvosym}
\definecolor{cvprblue}{rgb}{0.21,0.49,0.74}
\usepackage[pagebackref,breaklinks,colorlinks,allcolors=cvprblue]{hyperref}

\def\paperID{2818} 
\def\confName{CVPR}
\def\confYear{2026}

\title{AceTone: Bridging Words and Colors for Conditional Image Grading}

\author{
Tianren Ma$^{1,2}$\thanks{Work done during an internship at ByteDance.} \quad
Mingxiang Liao$^{2}$ \quad
Xijin Zhang$^{2}$ \quad
Qixiang Ye$^{1}$\thanks{Corresponding author. \Letter \  \tt{qxye@ucas.ac.cn}} \\[0.3ex]
$^{1}$ University of Chinese Academy of Sciences \quad
$^{2}$ ByteDance
}

\begin{document}
\maketitle

\begin{abstract}
Color affects how we interpret image style and emotion.
Previous color grading methods rely on patch-wise recoloring or fixed filter banks, struggling to generalize across creative intents or align with human aesthetic preferences.
In this study, we propose \textbf{AceTone}, the first approach that supports multimodal conditioned color grading within a unified framework.
AceTone formulates grading as a generative color transformation task, where a model directly produces 3D-LUTs conditioned on text prompts or reference images.
We develop a VQ-VAE-based tokenizer which compresses a $3\times32^3$ LUT vector to 64 discrete tokens with $\Delta \text{E}<2$ fidelity. 
We further build a large-scale dataset, AceTone-800K, and train a vision-language model to predict LUT tokens, followed by reinforcement learning to align outputs with perceptual fidelity and aesthetics.
Experiments show that AceTone achieves state-of-the-art performance on both text-guided and reference-guided grading tasks, improving LPIPS by up to 50\% over existing methods.
Human evaluations confirm that AceTone’s results are visually pleasing and stylistically coherent, demonstrating a new pathway toward language-driven, aesthetic-aligned color grading. Project Page: \url{github.com/martian422/AceTone}
\end{abstract}

\section{Introduction}
\label{sec:intro}

\begin{figure}[t]
  \centering
  \includegraphics[width=\linewidth]{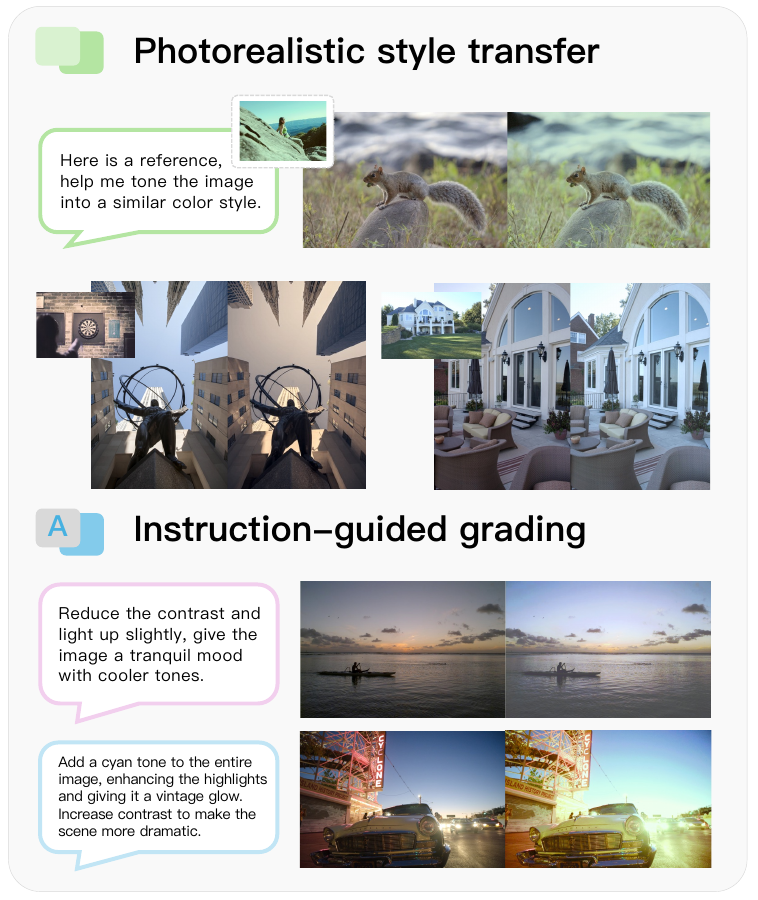}

   \caption{
   AceTone performs conditional color grading under two paradigms: reference-based (top) and instruction-based (bottom). It accurately captures subtle tonal characteristics, follows user intents, and produces visually coherent color adjustments.
   }
   \label{fig1}
\end{figure}

Before viewers recognize the content of a photograph, color already shapes their emotional and aesthetic impressions. 
Proper image grading (toning) not only guides visual attention but also enhances the conveyed style and emotion, improving the overall visual expressiveness.
However, existing automatic grading approaches mainly rely on recombining pre-defined filter libraries~\cite{zengLearningImageadaptive3D2020, condeNILUTConditionalNeural2024} or recoloring images patch by patch~\cite{luanDeepPhotoStyle2017,keNeuralPresetColor2023} with convolutional neural networks (CNNs). Although effective in some settings, these methods often fall short in expressiveness or efficiency.
In addition, for correlated tasks such as style transfer and instruction-guided grading, see Figure~\ref{fig1}, existing solutions are incompatible, and their segmented model designs puzzled the integration of a unified creative workflow.

A more profound challenge lies in the optimization objective of color grading models.
Existing approaches~\cite{keNeuralPresetColor2023, larchenkoColorTransferModulated2025,salut} typically introduce an adversarial design~\cite{goodfellowGenerativeAdversarialNets2014} to assess the similarity between the predicted result and the ground truth.
However, such a min–max adversarial paradigm is sensitive to initial hyper-parameters~\cite{meschederWhichTrainingMethods2018}, and suffers from mode collapse or unstable convergence~\cite{dhariwalDiffusionModelsBeat2021,hoDenoisingDiffusionProbabilistic2020}, making it difficult to scale training for consistent improvement. 
To address this, we postpone the alignment to a later stage. 
Instead of enforcing adversarial and perceptual loss during training, we first establish a stable likelihood-based generative model. 
We then incorporate color and aesthetic preferences through reinforcement learning (RL) with reward shaping, allowing the model to internalize aesthetic and perceptual constraints in a controllable and scalable manner.

Inspired by recent progress in aligning the generative model with preferences~\cite{ma2025hpsv3widespectrumhumanpreference,liuFlowGRPOTrainingFlow2025, wangUnifiedRewardModel2025}, we adapt a phased learning paradigm: (1) pre-train a model to learn the distribution of conditional color transformations using existing filters and reproducible grading examples, and (2) post-train the model with instruction patterns and aesthetic preference to learn what constitutes a good color grading.

To this end, we first introduce a tokenizer for 3-dimensional look up tables (3D-LUT), which uses a vector-quantize variational auto-encoder (VQ-VAE~\cite{razaviGeneratingDiverseHighFidelity2019}) to compress a continuous $3\times32^3$ LUT into 64 discrete tokens with high fidelity. 
Building upon this, we train a vision-language model (VLM) to generate LUT tokens conditioned on text or reference images, and construct a large-scale instruction dataset AceTone-800K with test suites. Finally, we integrate group relative policy optimization~\cite{guoDeepSeekR1IncentivizesReasoning2025} (GRPO) approach with aesthetic preference~\cite{deqa_score} and color similarity as composite rewards, guiding the model to produce visually pleasing and stylistically consistent results. 

Extensive experiments demonstrate that AceTone achieves state-of-the-art performance on both reference-based and text-guided color grading tasks. User studies further confirm its superiority in aligning with human visual preference and enhancing perceived beauty. The contributions of this study are summarized as follows:

\begin{itemize}
    \item \textbf{AceTone System}. We propose a novel and complete multimodal color grading framework that includes a high-fidelity LUT tokenizer and a VLM capable of generating color transformations directly from user instructions, marking the first generative color grading solution.
    \item \textbf{Generative and Reinforce Recipe}. We design a a stable and scalable learning strategy that aligns model outputs with aesthetics in color prediction.
    \item \textbf{Dataset and Benchmarks.} We build a large-scale dataset AceTone-800K and establish benchmarks with comprehensive meta data, providing valuable resources for learning semantic-aware color transformation.
\end{itemize}

\section{Background}

\noindent\textbf{Color Look Up Tables.}
A 3D-LUT defines a color mapping function $f: \mathbb{R}^3 \rightarrow \mathbb{R}^3$ that transforms input RGB values to a target appearance. 
In professional post-production workflows, LUTs are typically represented as cubes of size $3 \times N \times N \times N$, where each lattice vertex ($N^3$ in total) stores a normalized RGB triplet in the range $[0,1]^3$, describing the remapped color at a discrete sampling point. 
In short, a LUT can be regarded as a 3-channel volumetric tensor encoding color flow through RGB space, serving as a precise, interpretable \textit{color recipe} for image grading.

\noindent\textbf{Conditional Image Grading.}
Users intend to modify the color characteristics of an image while preserving its structural integrity. 
A major direction in this area is \textit{photorealistic style transfer} (PST)~\cite{salut}, where the goal is to mimic the color style of a reference image. 
Early approaches employ re-coloring methods that learn content-agnostic color representations and re-render target images via feed-forward networks~\cite{yooPhotorealisticStyleTransfer2019,hoDeepPresetBlending2021, keNeuralPresetColor2023,larchenkoColorTransferModulated2025}, which may incur prohibitive cost for high-resolution images. LUT-based image processing offers a more compact solution. 
Several methods~\cite{zengLearningImageadaptive3D2020,condeNILUTConditionalNeural2024, yang_adaint_2022, leonardis_image-adaptive_2025} model color grading as a weighted combination of pre-defined LUT presets. Recent work~\cite{salut} further proposes learnable LUT presets, but the applications are largely limited to log-space or specific domains.

In the context of \textit{instruction-guided grading} (IGG), some approaches~\cite{kwonCLIPstylerImageStyle2022,leeCLIPtoneUnsupervisedLearning2024} utilize CLIP to map text descriptions into color operations, but the input is confined to several words. Diffusion-based editing models~\cite{brooksInstructPix2PixLearningFollow2023,wuQwenImageTechnicalReport2025} allow text-conditioned recoloring by generating new images. 
Yet the latency and potential destructiveness make them unsuitable for modular or cascaded editing workflows.

\noindent\textbf{Aligning Generative Models with Preference.}
Recent advances~\cite{ma2025hpsv3widespectrumhumanpreference,liuFlowGRPOTrainingFlow2025, maConsolidatingReinforcementLearning2025} have shown remarkable income from aligning visual generation with subjective quality metrics via RL training. 
Inspired by these developments, our work leverages similar principles for color grading: 
rather than treating color mapping as a fixed regression task, we view it as a generative process that can be continuously refined through color and aesthetic rewards, enabling controllable and perceptually grounded color transformations.




%
%
%
\section{Method}
\begin{figure*}
    \centering
    \includegraphics[width=\linewidth]{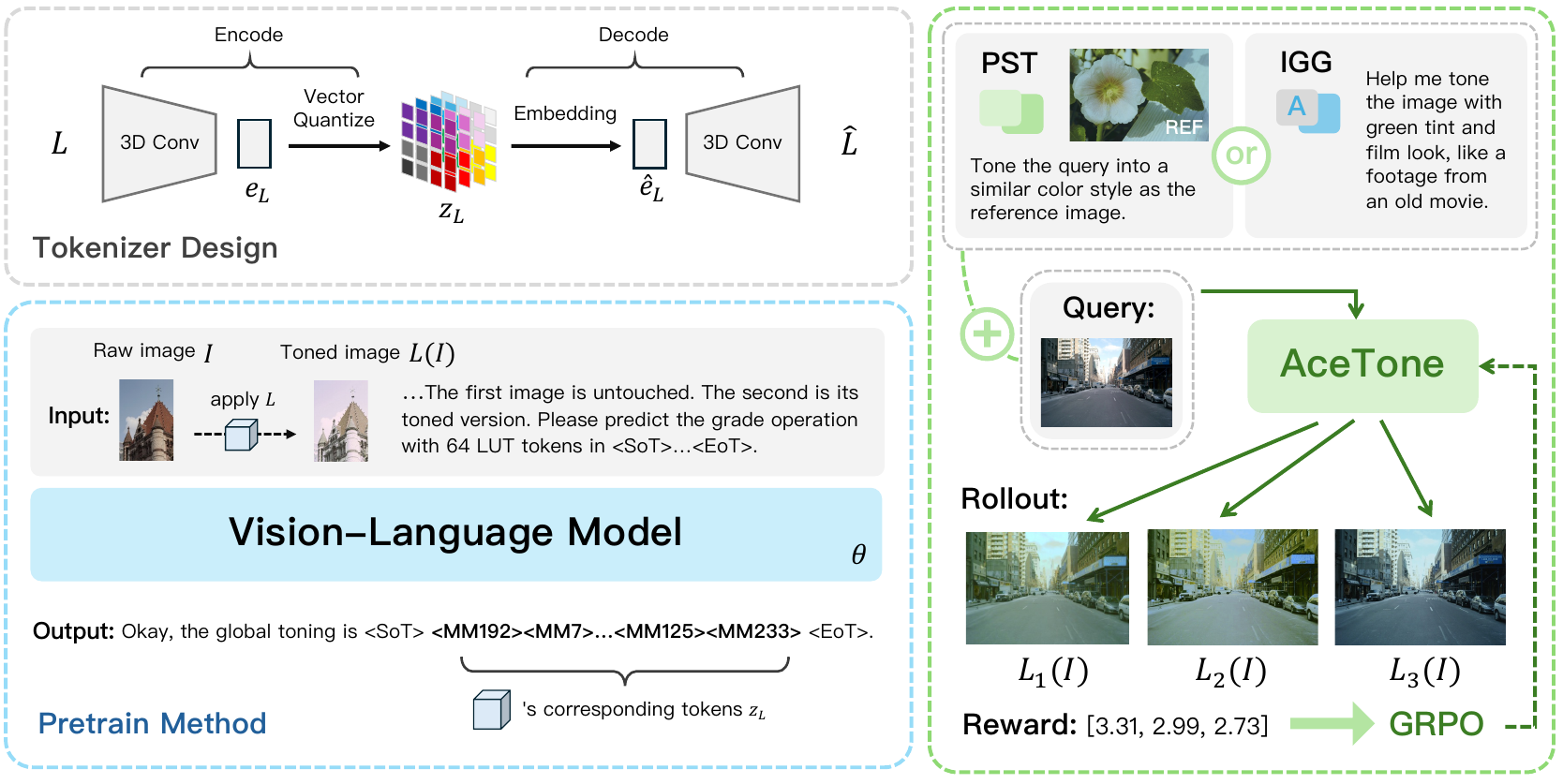}
    \caption{
    \textbf{Overview of the AceTone framework. }
    AceTone integrates a vector-quantized tokenizer to represent grading operations. Its phased training includes: (1) vision–language pretraining to learn tone prediction from image pairs, and (2) a post-train stage with SFT and RL (simplified in this figure). During RL training, multiple tone hypotheses are generated and optimized via group relative policy optimization (GRPO), enabling faithful and controllable tone manipulation aligned with user intent.}
    \label{fig:overview}
\end{figure*}

AceTone performs color grading by directly generating discrete LUT tokens that represent color transformations. As illustrated in Fig.~\ref{fig:overview}, the system is built to be fully compatible and efficient with VLMs, allowing multimodal conditioning from text prompts and reference images. Given a query image and multimodal instruction, the model learns to generate a structured sequence of LUT tokens. These predicted tokens are subsequently decoded into a high-fidelity 3D LUT, which can be directly applied to the input image as a \emph{non-destructive} color transformation. 

\subsection{LUT Tokenization}
To obtain a compact and discrete LUT representation, we design a 3D VQ-VAE tokenizer that compresses and reconstructs LUT volumes via vector quantization. Given a continuous LUT $L \in [0,1]^{3\times32\times32\times32}$, we first apply a cascaded learnable 3D convolutional encoder that progressively downsamples it to a latent tensor $e_L \in \mathbb{R}^{4\times4\times4\times D}$, where $D$ is the latent dimension. The latent features are then discretized by a vector-quantization layer $z_L = \text{Quantize}(e_L)$, forming a sequence of $T = 4^3$ tokens per LUT. The codebook of the quantization layer has $K$ entries, which indicates that each element of $ z_L$ is an integer in $\{0,1,...,K-1\}$.

During decoding, the quantized indices $z_L$ are embedded into continuous vector $\hat e_L\in \mathbb{R}^{4\times4\times4\times D}$ and upsampled to the original shape using three stages of learnable 3D convolutions, producing the reconstructed LUT $\hat{L}$. 

\noindent\textbf{Training Objective.}
The LUT tokenizer is optimized using a combination of  reconstruction loss and a vector-quantization commitment term. Given an input LUT $L$ and its reconstruction $\hat{L}$ produced by the decoder, the overall training objective is defined as
\begin{equation}
\mathcal{L} = \mathcal{L}_{\text{rec}} + \beta \, \mathcal{L}_{\text{commit}},
\end{equation}
where $\mathcal{L}_{\text{rec}} = \| L-\hat{L}  \|_2^2$ is the voxel-wise mean squared error (MSE) loss, and
\begin{equation}
\mathcal{L}_{\text{commit}} = \| e_L - \operatorname{sg}(\hat e_L) \|_2^2,
\end{equation}
denotes the commitment loss that constrains encoder outputs $e_L$ to stay close to their quantized counterparts $z_L$ while preventing gradients from flowing into the codebook via the stop-gradient operator $\operatorname{sg}(\cdot)$. 
The codebook entries are updated using exponential moving averages (EMA) of the encoder outputs, following VQ-VAE 2~\cite{razaviGeneratingDiverseHighFidelity2019}. 

%
%
%
\subsection{Next LUT Token Prediction}
Since the quantized LUT tokens represent a novel form of output for VLMs, a dedicated \textbf{generative pretraining} phase is required to adapt the model to valid LUT representations. The objective of this stage is to expose the model to a large number of image-LUT pairs via color-transfer tasks, \textit{i.e.,} original images and their graded counterparts, along with the corresponding LUTs. 

Formally, let the dataset be denoted as $\mathcal{D} = \{(I, L, c)\}$, where $I$ is the query image, and $c$ represents the text prompt. The LUT $L$ is quantized into a sequence of discrete tokens $z_L = \{z_1, z_2, \dots, z_T\}$ through the tokenizer. We provide the toned image $L(I)$ along the original $I$ to the VLM, and the objective is written as:
\begin{equation}
\mathcal{L}_{\text{gen}} = 
- \mathbb{E}_{(I, L, c) \sim \mathcal{D}} 
\sum_{t=1}^{T} 
\log p_\theta \big(z_t \mid z_{<t}, I, L(I), c \big),
\label{eq:gen_loss}
\end{equation}
where $p_\theta$ denotes the VLM parameterized by $\theta$. 

This objective trains the model to auto-regressively predict LUT tokens conditioned on both visual and textual contexts, aligning the generative capacity of the VLM with the structured color transformation space.

\subsection{Post-training}
\paragraph{Supervised Fine-Tuning (SFT).}
After establishing the VLM's basic ability to predict LUT tokens through generative pretraining, we further fine-tune the model to adapt to realistic use cases, including \textit{photorealistic style transfer} (PST) and \textit{instruction-guided grading} (IGG). This stage enables the model to generalize from synthetic supervision to real-world, context-aware color grading tasks.

For the task of mimicking the color style of a reference image, we construct paired data samples $\mathcal{D}_{\text{style}}=\{(I_1, I_2, L, \phi, c)\}$, where $I_1$ and $I_2$ are two randomly sampled images, $\phi$ is a randomly applied low-intensity color perturbation for data augmentation. The model is required to predict the quantized LUT tokens $z_L$ based on the reference image $L(\phi(I_2))$ and perturbed query image $\phi(I_1)$. 
Formally, the training objective is defined as
\begin{equation}
\begin{aligned}
\mathcal{L}_{\text{style}}  = 
& - \mathbb{E}_{(I_1, I_2, L, \phi, c) \sim \mathcal{D}_{\text{style}}} \\
&\sum_{t=1}^{T}
\log p_\theta \big(z_t \mid z_{<t}, \phi(I_1), L(\phi(I_2)), c \big).
\label{eq:style_loss}
\end{aligned}
\end{equation}

To enable text-driven color grading, we construct an instruction dataset 
$\mathcal{D}_{\text{inst}} = \{(I, L,c_L)\}$, where $c_L$ denotes a textual instruction describing the desired color adjustment. The instructions are annotated by comparing the ungraded image $I$ and its graded version $L(I)$ using the Qwen2.5-VL-32B model, which infers concise human-like editing descriptions (see Section~\ref{sec:dataset} for details). 
The objective encourages the model to predict the LUT token sequence conditioned on the query image and textual instruction, as
\begin{equation}
\mathcal{L}_{\text{inst}} = 
- \mathbb{E}_{( I, L,c_L) \sim \mathcal{D}_{\text{inst}}}
\sum_{t=1}^{T}
\log p_\theta \big(z_t \mid z_{<t},I,c_L \big).
\label{eq:inst_loss}
\end{equation}
Through joint optimization of $\mathcal{L}_{\text{style}}$ and $\mathcal{L}_{\text{inst}}$, the model learns to interpret multimodal inputs, achieving primarily controllable and photorealistic color grading.

\paragraph{Reinforcing AceTone with GRPO.}
After fine-tuning, the resulted model learns to follow instructions and imitate reference styles. However, it does not guarantee that the generated results are aesthetically pleasing or faithful to human intent. To further improve the stability and practical usefulness of its outputs, we introduce a reinforcement learning stage that explicitly aligns the model’s color grading behavior with human preference and color similarity.

Let $\textbf{c}$ be a multimodal prompt, which may instruct the model to grade the query image $I$ based on reference image or text, the model predict a set of LUT tokens. For the predicted LUT $L_{\text{pred}}$, we get the graded image $L_{\text{pred}}(I)$ and compute complementary reward functions:
\begin{itemize}
    \item \textbf{Color similarity} $r_{\text{color}}(\cdot,I_{\text{gt}})$: measures the color consistency between the graded image and the ground-truth using perceptual color metric $\Delta \text{E}(I_{\text{gt}}, L_{\text{pred}}(I))$. The reward is calculated as $\frac{1}{\max(2, \,\Delta \text{E}) - 1}$, which indicates that the maximum value is given when $\Delta \text{E}<2$.
    \item \textbf{Aesthetic quality} $r_{\text{aes}}(\cdot)$: evaluates the visual appeal of the graded image using a pretrained aesthetic assessment model, which gives a continuous score $\text{Aes}(L_{\text{pred}}(I))$ in $[0,5]$, and then scaled to $[0,1]$ as the aesthetic reward. 
\end{itemize}

GRPO~\cite{guoDeepSeekR1IncentivizesReasoning2025} samples a group of $G$ responses $\{o_1, o_2, \dots,o_G\}$ from the policy $\pi_\theta$. For rollout $o_i$, the reward system gives an action value $r_i$. We omit the clip operation for clarity. With $\rho_i^k$ as the importance of the $k$-th token in $o_i$, the advantage $A_i$ and reward component $R(\theta)$ is calculated as:
\begin{equation}\label{adv}
    A_i = \frac{r_i - \text{mean}(\{r_j\}_{j=1}^G)}{\text{std}(\{r_j\}_{j=1}^G)}, R(\theta) =  
    \frac{1}{G}\sum_{i=1}^G \frac{1}{|o_i|} \sum_{k=1}^{|o_i|}
     \rho_i^k A_i
\end{equation}
The GRPO objective is expressed as
\begin{equation}\label{grpo}
\max_{\theta} \,\mathbb{E}_{\textbf{c}\sim\mathcal{D}}\Big[R(\theta) - \beta 
 \mathbb{D}_{\text{KL}} \big[ \pi_\theta(\cdot) \,\|\, \pi_{\text{ref}}(\cdot) \big] \Big],
\end{equation}
where $\beta$ regulates the strength of the KL regularization. 

A detailed description is provided in Alg.~\ref{algo-grpo}.
\begin{algorithm}[t]
\caption{GRPO for AceTone}
\label{algo-grpo}
\begin{algorithmic}[1]
\Require Reference model $\pi_{\text{ref}}$, prompt-image dataset $\mathcal{D}$, number of completions per prompt $G$. 
\State Initialize policy $\pi_\theta \gets \pi_{\text{ref}}$
\While{not converged}
    \State Sample prompt-image pair $(\mathbf{c}, I, I_{\text{gt}}) \sim \mathcal{D}$
    \State Sample $G$ completions $o_i \sim \pi_{\theta}(\cdot \mid \mathbf{c}, I)$
    \State For each $o_i$, decode it into LUT $L_i$
    \State Get reward $r_i \gets r_{\text{color}}\big(L_i(I),I_{\text{gt}}\big)+r_{\text{aes}}\big(L_i(I)\big)$
    \State Compute objective with Eq.~\ref{grpo} and update $\pi_\theta$ 
\EndWhile
\State \Return $\pi_\theta$
\end{algorithmic}
\end{algorithm}

\section{Dataset}
\label{sec:dataset}

\subsection{Dataset Collection}
To support large-scale training of AceTone, we construct a comprehensive dataset that combines diverse LUT sources, expert edits, and curated instruction annotations. 

\noindent\textbf{LUT libraries.}
We curate a large and diverse LUT corpus as the foundation of our training data.  

(1) Filter Library.
We collected approximately 10{,}000 licensed LUT filters in \texttt{.cube}. To ensure consistency, all LUTs are resampled into the 32-bit ($3\times32 \times 32 \times 32$) format. This library exhibits wide stylistic diversity and is used primarily to learn general LUT representations.

(2) Expert Library.  
PPR-10K~\cite{liangPPR10KLargeScalePortrait2021} provides professional color adjustments from three experts on 10k raw images. These adjustments are originally stored in Adobe Lightroom templates. We convert them into applicable 32-bit LUTs (about 34{,}000 total) via the automated export pipeline. These LUTs reflect artistic and subtle adjustments, helping the model learn delicate post-processing intentions.

(3) Fuse Library.
To reduce redundancy and improve representativeness, we use PCA~\cite{mackiewiczPrincipalComponentsAnalysis1993} and K-means to cluster the union of the above two libraries into a compact set of 8{,}192 LUTs. This library serves as the basis for instruction annotation and evaluation.


\noindent\textbf{Image datasets.}
We use MS-COCO~\cite{linMicrosoftCOCOCommon2014}, Adobe-5K~\cite{adobefivek}, and PPR-10K~\cite{liangPPR10KLargeScalePortrait2021} as our primary image sources. MS-COCO offers real-world stylistic diversity, while Adobe-5K and PPR-10K mainly contain ungraded images, making them suitable for post-training stages.

\subsection{Dataset Curation}
\noindent\textbf{LUT Tokenizer Training.}
The tokenizer is trained exclusively on the filter library, while evaluation samples are drawn from the expert library to assess generalizability.

\noindent\textbf{Generative Pretraining.}
Following Neural Preset~\cite{keNeuralPresetColor2023}, we use MS-COCO as the image corpus. 
Each image is randomly paired with 64 LUTs sampled from the filter library and a pre-defined prompt, forming $(I,L,c)$ triplets.

\noindent\textbf{Supervised Fine-tuning.}
We use Adobe-5K and PPR-10K as the image sources. 
For PST, LUTs are sampled from the fuse library and applied randomly to create paired supervision. For fair evaluation, we hold out 1024 samples with \textbf{non-overlapping LUTs and images} as the $\text{AceTone-Bench}[\text{Transfer}]$ set. The construction process follows the same protocol as PST-50~\cite{salut}. 

To build data for IGG, we employ the Qwen2.5-VL-32B~\cite{baiQwen25VLTechnicalReport2025} model to generate editing intentions. 
We sample 300K image–LUT pairs $(I,L)$, the VLM produces three alternative editing instructions for each pair. We manually inspected 512 randomly sampled annotations from raw data and found that $\sim$10\% lacked clear color directionality. Such samples were rejected via LLM auto-detection, resulting a corpus of $\sim$800K automatically annotated tuples $(I,\, L(I), L,c_L)$.
We reserve 128 high-quality examples with strict human check, forming the $\text{AceTone-Bench}[\text{Instruct}]$ set.

To our knowledge, this is the most comprehensive dataset and benchmark curation for conditional color grading (compared in Table~\ref{tab:bench}), with detailed and reproducible meta data. 

\begin{table}[h]
\centering
\setlength{\tabcolsep}{0.09cm}
\caption{Comparison of existing color grading benchmarks.}
\begin{tabular}{lccc}
\toprule
Benchmarks & Task & \#Samples & Ground Truth\\
\midrule
DPST~\cite{luanDeepPhotoStyle2017} & Transfer & 60 & none \\
PST-50~\cite{salut} & Transfer & 50 & image\\
MMArt-Bench~\cite{linJarvisArtLiberatingHuman2025} & Instruct & 200 & cropped image \\
\midrule
\textbf{AceTone-Bench} & 
\begin{tabular}{@{}c@{}}Transfer\\ Instruct\end{tabular} &
\begin{tabular}{@{}c@{}}1024 \\128 \end{tabular} &
\begin{tabular}{@{}c@{}}image \& LUT\\image \& LUT \end{tabular}\\
\bottomrule
\end{tabular}
\label{tab:bench}
\end{table}

\noindent\textbf{Reinforcement Learning.}
For RL training, we sample 30K $(\textbf{c},I, I_{\text{gt}})$ pairs for PST and IGG task respectively. These pairs are randomly extracted from the SFT data. The aim of this stage is to shape the model's behavior via rewards, which demands the data pairs to be diverse. Please refer to Section~\ref{sec:ablation} for data distribution's effect on results.

\section{Experiment}\label{sec:experiments}
Building upon the datasets in Section~\ref{sec:dataset}, we present the training protocol, evaluation metrics, and experimental results for AceTone. 
Our experiments include both qualitative visualizations and quantitative comparisons against existing methods. 
Additionally, we conduct ablation studies to analyze several choices for training AceTone.
\subsection{Implementation Detail}
We first train a general-purpose LUT tokenizer that encodes and decodes LUTs in a quantized token form. To enhance robustness, we perform data augmentations on LUT vectors, including random gamma correction, contrast, exposure jitter, and smoothed Gaussian noise. The codebook size $K$ is set to 256. Training is conducted using Adam optimizer~\cite{kingmaAdamMethodStochastic2014} with a learning rate of $2\times10^{-4}$ and a batch size of $64$. We set a constant commitment weight $\beta = 0.25$. The tokenizer is trained for $500$ epochs on 8 GPUs, totaling about $7$ hours.

For the VLM, we initialize from Qwen2.5-VL-3B~\cite{baiQwen25VLTechnicalReport2025}, and extend its vocabulary to include discrete LUT token entries, enabling the model to directly predict logits over LUT tokens. The visual encoder is frozen, while we tune the MLP connector and language model.
The model is trained for two epochs on the generative pre-training, and tuned for one epoch for the following stages. The training process takes about $3$ days with 8 GPUs.

\subsection{Evaluation Metric}

We evaluate AceTone using both aesthetic and quantitative metrics: \textbf{Aesthetic}: we use DeQA~\cite{deqa_score}, an aesthetic-aware expert model to automatically evaluate the quality of photorealistic images. \textbf{PSNR} and \textbf{LPIPS}~\cite{zhangUnreasonableEffectivenessDeep2018}: measure pixel-wise and perceptual similarity between prediction and ground-truth images. $\bf \Delta \bf E$: quantifies the color difference with CIEDE2000 standard~\cite{sharmaCIEDE2000ColordifferenceFormula2005}, specialized for assessing the deviation of color tones between the prediction and ground-truth images.
For each task, we report the average scores on the corresponding evaluation splits from PST-50~\cite{salut}, AceTone-Bench$[\text{Transfer}]$ and AceTone-Bench$[\text{Instruct}]$.  

\subsection{Comparative Evaluation}\label{sec:results}

\begin{table*}[t]
\centering
\small
\caption{\textbf{Comparison on photorealistic style transfer.} Aes. stands for the aesthetic score given by DeQA. $^\dagger$: Neural Preset has not been open-sourced, and is only available as a mobile app without support for batch process, therefore we manually evaluate it on PST-50.}
\begin{tabular}{lccccccccc}
\toprule
\multirow{2}{*}{Method} & \multirow{2}{*}{Venue} & \multicolumn{4}{c}{PST-50 (50 samples)} & \multicolumn{4}{c}{AceTone-Bench$[\text{Transfer}]$ (1024 samples)} \\
\cmidrule(lr){3-6} \cmidrule(lr){7-10}
 & & Aes. $\uparrow$ & PSNR $\uparrow$ & LPIPS $\downarrow$ & $\Delta \text{E} \downarrow$ & Aes. $\uparrow$ & PSNR $\uparrow$ & LPIPS $\downarrow$ & $\Delta \text{E} \downarrow$  \\
\midrule
Neural Preset$^\dagger$~\cite{keNeuralPresetColor2023}  &CVPR-23 & 3.03 & 21.24 & 0.15 & 9.57 & -- & -- & -- & -- \\
WCT$^2$~\cite{yooPhotorealisticStyleTransfer2019}  &ICCV-19 & 2.95 & 19.62 & 0.18 & 10.91  & 2.69 & 15.12 & 0.32 & 18.23 \\
ModFlow~\cite{larchenkoColorTransferModulated2025}  &AAAI-25& 3.08 & 20.13 & 0.16 & 10.62  & 2.96 & 15.61 & 0.31 & 17.79 \\
SA-LUT~\cite{salut}  &ICCV-25& 3.07 & 21.64 & 0.16 & 9.01  & 3.33 & 18.14 & 0.22 & 13.10 \\ 
\midrule
\textbf{AceTone (ours) } & & \textbf{3.29} & \textbf{24.26} & \textbf{0.09} & \textbf{7.26} & \textbf{3.57} & \textbf{22.49} & \textbf{0.11} & \textbf{8.98} \\
\bottomrule
\end{tabular}
\label{tab:transfer}
\end{table*}

\begin{table}[h]
\centering
\small
\setlength{\tabcolsep}{0.09cm}
\caption{\textbf{Comparison on instruction-guided grading.} Arc. refers to the model architecture, \textit{e.g.}, diffusion, parameter-based software agent, or color transformation with LUTs.}
\begin{tabular}{lccccc}
\toprule
Method & Arc. & Aes. $\uparrow$ & LPIPS $\downarrow$ & Rank & Time\\
\midrule
InstructPix2Pix~\cite{brooksInstructPix2PixLearningFollow2023} & Diff. & 2.97 & 0.44 & 3.68 & 2s \\
Qwen-Image-Edit~\cite{wuQwenImageTechnicalReport2025} & Diff. & 3.19 & 0.33 & 2.37 & 2min \\
JarvisArt~\cite{linJarvisArtLiberatingHuman2025} & Agent & 3.17 & 0.31 & 2.72 & 40s\\
\midrule
\textbf{AceTone (ours)} & LUT & \textbf{3.54} & \textbf{0.22} & \textbf{1.43} & \textbf{1.2s} \\
\bottomrule
\end{tabular}
\label{tab:instruct}
\end{table}

\begin{figure}[t]
  \centering
  \includegraphics[width=\linewidth]{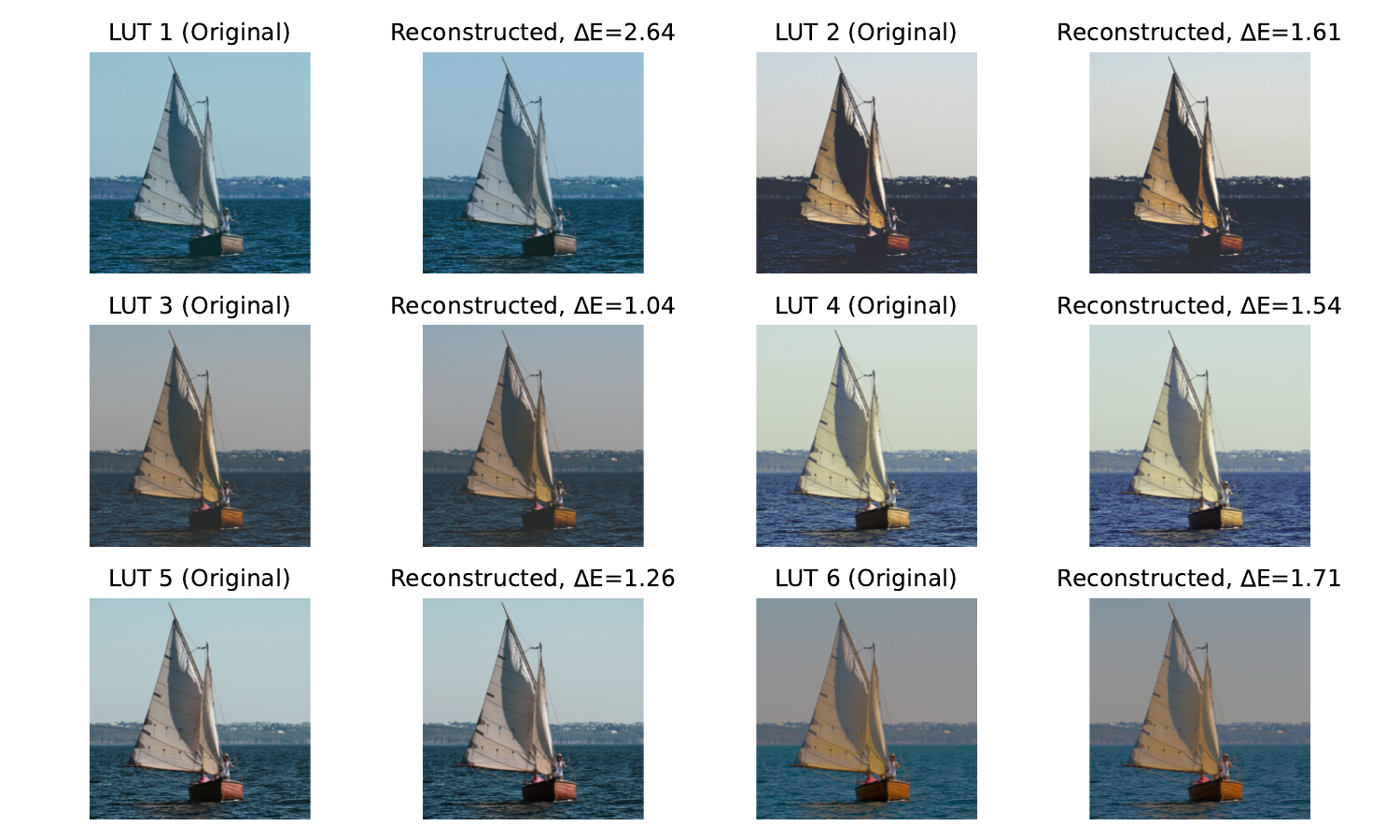}

   \caption{
   \textbf{Visualization of the tokenizer's compression quality.} We apply the original and reconstructed LUT to the image, and report the corresponding $\Delta \text{E}$ metric.
   }
   \label{fig:tokenizer}
\end{figure}

\noindent\textbf{Tokenizer Evaluation.}
Our LUT tokenizer achieves an average PSNR of 37.5\,dB on a held-out set of 1024 test LUTs, establishing a strong foundation for subsequent generative modeling. 
To further evaluate perceptual fidelity, we apply both the original and reconstructed LUTs to natural images from the Adobe-5K dataset and measure their color deviation. 
The resulting $\Delta \text{E} = 1.38$ indicates that the reconstruction difference is virtually imperceptible (see Figure~\ref{fig:tokenizer} for qualitative examples). 
Notably, the entire tokenizer contains only $\sim$4M parameters, making it compact and easily deployable on modern devices.

One continuous alternative \cite{zehtabEfficientNeuralNetwork2024} achieves a 99.87\% compression ratio with $\Delta \text{E} = 3.11$. 
As the first work to introduce a \textit{discretized} compression scheme for 3D-LUTs, our tokenizer attains a \textbf{99.98\%} ratio while maintaining a significantly lower color error of $\Delta \text{E} = \textbf{1.38}$, demonstrating its high-fidelity quantization and effective LUT representation.

\noindent\textbf{Quantitative Comparison.}
Table~\ref{tab:transfer} summarizes the quantitative evaluation on PST. Across all major metrics, AceTone consistently outperforms existing approaches by a significant margin. AceTone's performance on the established PST-50~\cite{salut} benchmark is achieved in a \textbf{\textit{strictly zero-shot}} manner, which means neither the toning methods nor images with similar styles (frames from raw video clips) are ever encountered during training.
In addition, the proposed AceTone-Bench$[\text{Transfer}]$ establishes a challenging test suite with a wide spectrum of styles, contributing to a comprehensive assessment of PST models.

Comparisons on AceTone-Bench$[\text{Instruct}]$ are presented in Table~\ref{tab:instruct}. 
We include two representative diffusion-based models and one recent agent-based 7B VLM for reference. 
Beyond standard quantitative metrics, we further employ Gemini-2.5-Pro~\cite{Gemini25Pro} as a preference evaluator by prompting it to rank results given the instruction, the unedited image, and four graded candidates. 
Improvements in LPIPS and averaged rank demonstrate that AceTone not only reproduces target color distributions more faithfully but also generates visually pleasing results that better align with aesthetic judgments. 
Among all models supporting text-based control, AceTone requires only $\sim$1\,s per request, matching the fastest method while achieving the best performance. 
When integrated with vLLM~\cite{kwon2023efficientvllm}, inference latency can be further reduced to under 300\,ms.

%

\subsection{Qualitative Result}

\begin{figure*}
    \centering
    \includegraphics[width=\linewidth]{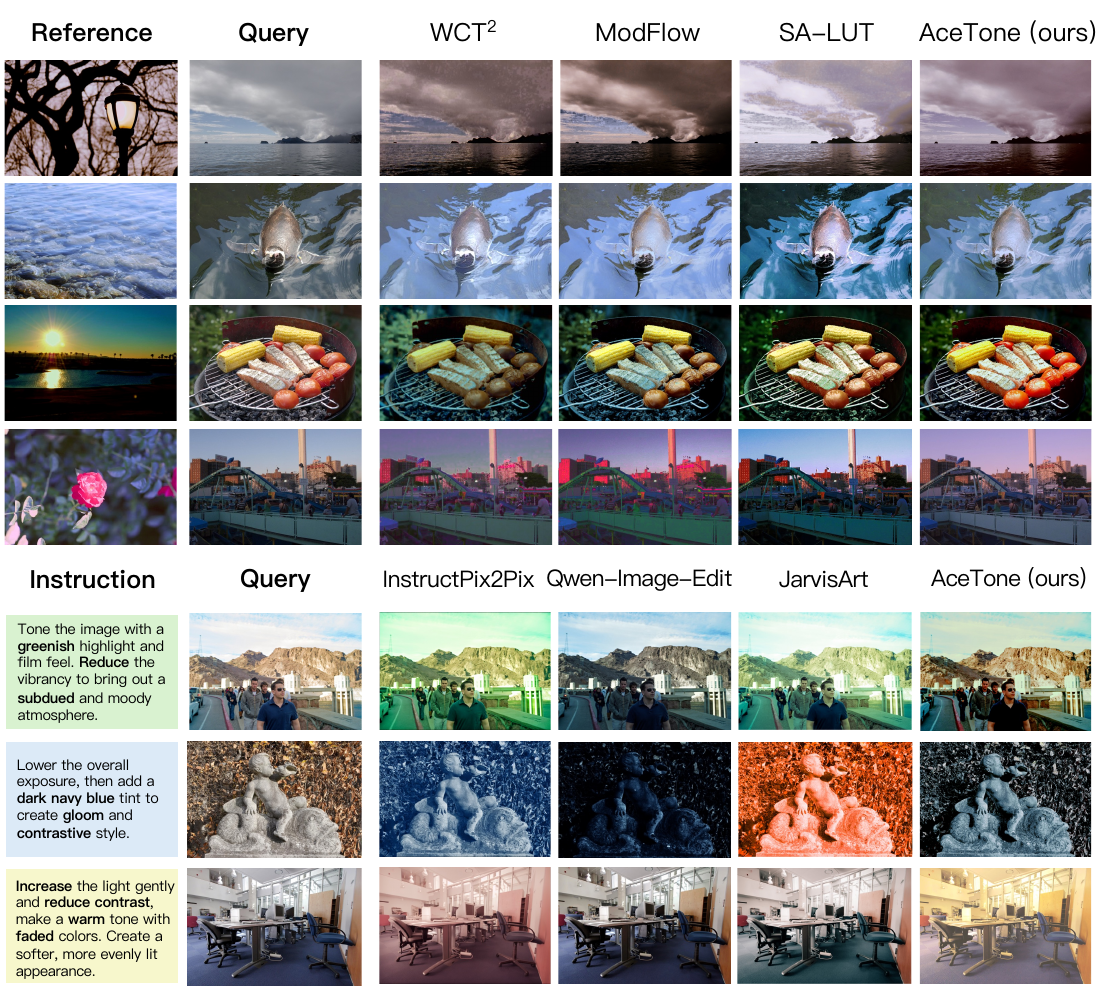}
    \caption{
    \textbf{A Qualitative Visualization of AceTone.}
    Top: Style transfer comparison. AceTone effectively captures the target color style while avoiding color banding or unnatural hue shifts. Bottom: Instruction-guided grading comparison (from AceTone-Bench $[\text{Instruct}]$). AceTone faithfully follows user intent and produces visually appealing adjustments. Best viewed in color and zoomed in for full details. }
    \label{fig:qualitative}
\end{figure*}

Figure~\ref{fig:qualitative} presents a qualitative evaluation of AceTone on both photorealistic style transfer (top) and instruction-guided grading (bottom) tasks. In the style transfer setting, AceTone effectively captures the color style from the reference image while preserving the natural color distribution of the query image. Compared to prior methods, AceTone avoids common artifacts such as color banding, over-saturation, and unnatural hue shifts, resulting in visually coherent and aesthetically balanced outputs. In the instruction-guided grading task, AceTone consistently adheres to user-specified grading directions (marked in \textbf{bold}), such as introducing greenish film tones, applying dark navy tints, or creating warm and soft appearances, while maintaining fine structural details and tonal consistency. These results demonstrate that AceTone faithfully interprets user intent and produces visually appealing color adjustments that outperform existing baselines.


\subsection{Ablative Study}\label{sec:ablation}

\begin{figure}[t]
  \centering
  \includegraphics[width=\linewidth]{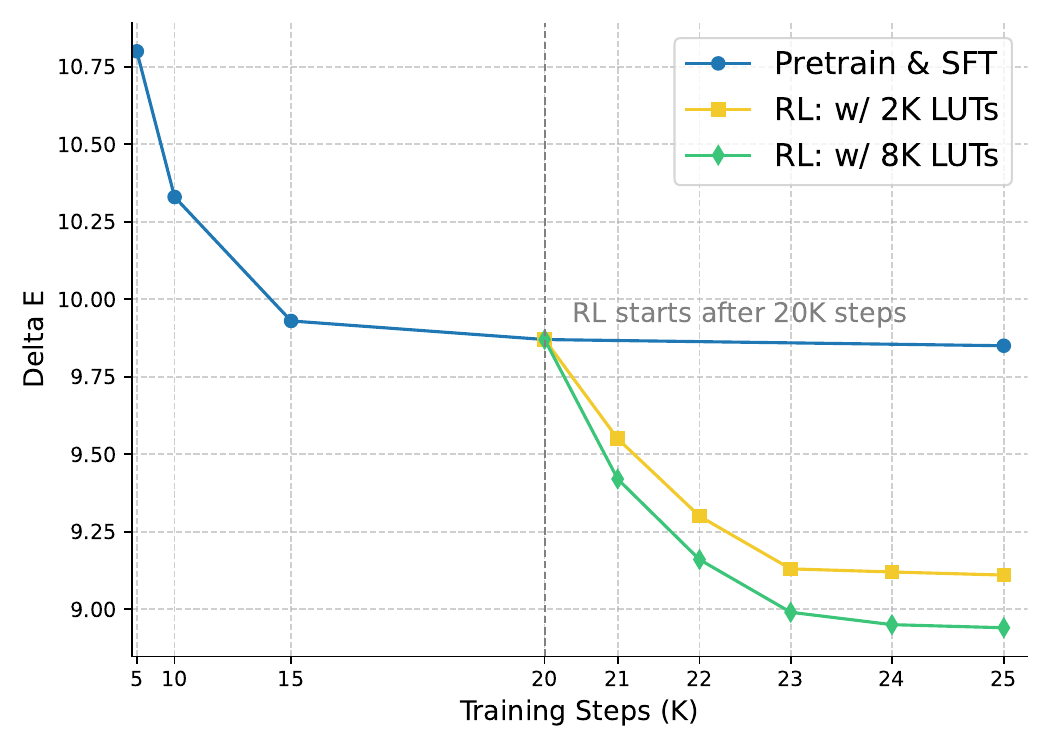}

   \caption{
   \textbf{Performance Curve during Training.}
   }
   \label{fig:rl}
\end{figure}

\noindent\textbf{RL Effectiveness.} 
We evaluate AceTone on AceTone-Bench$[\text{Transfer}]$ across different training stages to monitor performance evolution, as illustrated in Figure~\ref{fig:rl}. 
While the performance tends to saturate at the late stage of SFT, reinforcement learning brings a clear and consistent improvement—note that AceTone already surpasses all existing methods after SFT. 
This confirms the effectiveness of GRPO in refining model behavior through reward-driven optimization. 
We further experiment with different RL data settings by varying the size of the LUT pool used for data construction. 
Models trained under a more challenging setup (with 8K LUTs for curation) exhibit better generalization, highlighting the importance of data diversity during RL training.

\noindent\textbf{Reward Design.} 
Although color-related performance can be assessed using various metrics, we observe that PSNR, LPIPS, and $\Delta \text{E}$ follow nearly identical trends throughout training. 
In practice, the proposed color similarity reward already provides a sufficiently informative signal for RL optimization, and adding additional metric-aligned rewards yields no further gain, as shown in Table~\ref{tab:ablation}. 
Therefore, our final configuration retains only the color similarity reward alongside the aesthetic reward to balance perceptual fidelity and stylistic preference.

\begin{table}[h]
\centering
\setlength{\tabcolsep}{0.09cm}
\caption{\textbf{Ablation on Reward Designs.}}
\begin{tabular}{lcccc}
\toprule
Reward Metrics  & PSNR $\uparrow$ & LPIPS $\downarrow$ & $\Delta \text{E} \downarrow$ & Aes. $\uparrow$\\
\midrule
$\Delta \text{E}$ & 22.51 & 0.11 & 8.93 & 3.29 \\
$\Delta \text{E}$ \& LPIPS & 22.24 & 0.10 & 9.05 & 3.28\\
$\Delta \text{E}$ \& PSNR & 22.58 & 0.11 & 9.01 & 3.31\\
$\Delta \text{E}$ \& Aes. & 22.49 & 0.11 & 8.98 & 3.57\\
\bottomrule
\end{tabular}
\label{tab:ablation}
\end{table}

\section{Discussion}

\noindent\textbf{User Study.}
We further conduct a user preference study with 20 participants, comparing outputs from AceTone and competing methods under 15 tests for each task. We report the average preference, which indicates the percentage of being nominated as the best version, and the average rank given by users. As shown in Table~\ref{tab:user}, participants consistently favor AceTone’s results for their stylistic coherence, and faithful interpretation of prompts, confirming the effectiveness of our RL-enhanced generative formulation for conditional color grading.

\begin{table}[h]
\centering
\small
\setlength{\tabcolsep}{0.09cm}
\caption{\textbf{User Study.} The correlated statistics are: Friedman $\chi ^2 (2)=58.14, p< 0.001$ and inter-rater agreement $\kappa=0.213$.}
\begin{tabular}{lcccc}
\toprule
Method  & Task & Pref. (\%) & Rank \\
\midrule
Neural Preset~\cite{keNeuralPresetColor2023} & \multirow{3}{*}{PST} & 30.1 & 2.03  \\
SA-LUT~\cite{salut} & & 28.5 & 2.22 \\
AceTone (ours) & & \textbf{41.4} & \textbf{1.75} \\
\midrule
JarvisArt~\cite{linJarvisArtLiberatingHuman2025} & \multirow{3}{*}{IGG} & 18.3 & 2.34  \\
Qwen-Image-Edit~\cite{wuQwenImageTechnicalReport2025} & & 34.8 & 1.97 \\
AceTone (ours) & & \textbf{46.9} & \textbf{1.69} \\
\bottomrule
\end{tabular}
\label{tab:user}
\end{table}

\noindent\textbf{Application.}
Beyond its research contribution, AceTone opens up a wide range of practical applications in visual content creation and editing. As a multimodal prompt-driven grading system, it can serve as a universal color assistant across photo and video platforms, enabling users to express complex tonal intentions through text prompts and reference images. Due to its LUT-based nature, AceTone can also be seamlessly integrated into existing post-production pipelines, offering \textbf{\textit{non-destructive}} color adjustments for professional tools like desktop or mobile editing apps. 
Besides, AceTone can function as a powerful data generator: by producing high-quality, controllable color transformations, it can augment datasets for aesthetic learning, domain adaptation, or relighting tasks. 

\noindent\textbf{Limitation.}
We observe failure modes under extreme illumination or highly stylized scenes, where global LUTs cannot fully capture local lighting or texture-dependent color shifts. VLM's perception to color may be flawed~\cite{liang_colorbench_2025}, leading to less informative annotations.
Instruction-guided grading may also misinterpret ambiguous or culturally nuanced terms such as “nostalgic tone” or “soft cinematic.” 
Besides, due to time and cost constraints, the user study was distributed online. All participants used Macbooks or external wide-gamut displays, but we did not enforce calibration or vision screening.

\section{Conclusion}
In this work, we introduce AceTone, a generative system that bridges language and color for conditional image grading. AceTone directly produces 3D-LUT–based color transformations, supported by a high-fidelity tokenizer and a dedicated training scheme. The proposed AceTone-800K and benchmark further contribute to the research of semantic-aware color transformation tasks. Experiments demonstrate AceTone's state-of-the-art performance on both text- and reference-guided grading tasks.

We believe AceTone offers a new direction for multimodal prompt-driven visual post-processing, where controllable and human-aligned color transformation serves as a bridge between visual semantics and creative intent. Future work may extend this paradigm toward interactive grading, video color dynamics, and the broader study of computational aesthetics.

{
    \small
    \bibliographystyle{ieeenat_fullname}
    \bibliography{main}
}


\end{document}